\documentclass[journal]{IEEEtran}
\ifCLASSINFOpdf
\else
\fi
\usepackage{amsmath}
\usepackage{graphicx}

\begin{document}
%
\title{Let CONAN tell you a story: Procedural quest generation}
%
%
%

\author{Vincent~Breault, S\'ebastien~Ouellet, Jim~Davies

\thanks{Vincent Breault and S\'ebastien Ouellet were with the Institute of Cognitive Science, Carleton University, Ottawa, Canada, e-mail: breault.vincent@gmail.com.}
\thanks{Jim Davies is with the Institute of Cognitive Science, Carleton University, Ottawa, Canada.}}

\markboth{IEEE TRANSACTIONS ON COMPUTATIONAL INTELLIGENCE AND AI IN GAMES}%
{}
%



\maketitle

\begin{abstract}
This work proposes an engine for the Creation Of Novel Adventure Narrative (CONAN), which is a procedural quest generator. It uses a planning approach to story generation. The engine is tested on its ability to create quests, which are sets of actions that must be performed in order to achieve a certain goal, usually for a reward. The engine takes in a world description represented as a set of facts, including characters, locations, and items, and generates quests according to the state of the world and the preferences of the characters. We evaluate quests through the classification of the motivations behind the quests, based on the sequences of actions required to complete the quests. We also compare different world descriptions and analyze the difference in motivations for the quests produced by the engine. Compared against human structural quest analysis, the current engine was found to be able to replicate the quest structures found in commercial video game quests.
\end{abstract}


%
\IEEEpeerreviewmaketitle

\section{Introduction}

  The creation of media content has always been the domain of humans, be it for movies, music or video games. With advancement in computer technology and research, the creation of such content has seen a slight shift from the human authored to automatic computer generation. Using algorithms to procedurally create media can effectively alleviate some of the burden from artists when creating a new piece. 
    
\subsection{Procedural Generation in Games}
Procedural Content Generation for Games (PCG-G) is the use of computers algorithms to generate game content, determine if it is interesting, and select the best ones on behalf of the players.\cite{hendrikx2013procedural}
    
This type of generation becomes quite useful when trying to produce content for an industry that is more and more demanding in terms of content \cite{hendrikx2013procedural}. For instance, in the current market, game development costs are extremely high as the demand for highly complex games requires the work of many artists and many hours to be met. For instance, the Massively Multiplayer Online Role Playing Game (MMORPG) World of Warcraft has a total of 30,000 items, 5,300 creatures with which to interact and 7600 quests and has an estimated budget of twenty to one hundred and fifty million dollars for a single game \cite{hendrikx2013procedural}. An engine capable of offloading this task by automatically generating such content would be invaluable to the industry, as it would greatly reduce development costs by reducing the amount of work that has to be done by humans.

Procedural generation (PG) in video games is currently used in a variety of different subfields to generate content. Indeed, where artists have previously been used to create everything, some of the work has been offloaded to automation for various kinds of design, such as graphics (e.g., textures in \cite{lefebvre2003pattern}) or vegetation, through such programs as SpeedTree \cite{de2009tools}. The creation of environment and maps has also seen its share of procedural generation algorithms for the generation of buildings \cite{martin2010evolving} \cite{kelly2007citygen}, road networks \cite{sun2002template} and maps \cite{hartsook2011toward,togelius2010towards}. PG can also be observed in the automatic or custom creation of more abstract elements such as ships or weaponry \cite{hastings2009automatic}, Non-Player Character (NPC) behaviour or other such system behaviour \cite{orkin2005agent}.
    
\subsection{Narrative generation}

  Another area where PG is used to generate abstract elements is narrative generation. Characters in games give “quests'' to the player character (PC) \cite{doran2010towards}. A quest is a set of actions that must be performed in order to achieve a certain goal, usually for a reward. They are usually provided by non-player characters within the game and are embedded in a piece of narrative that makes the sequence of actions make sense given the NPC and the current world state.  Some quests are simple, but some are complex. For example, the quest ``Cure for Lempeck Hargrin'' that Doran uses for comparison has 27 steps, while others might have only a few \cite{doran2010towards}. 
    
\subsubsection{Planning}    
    Given the structural similarities between the outputs of AI planning agents and quests, planning has seen a lot of use in the PG of stories. People have used planning and machine learning to modify previously human authored stories \cite{li2010offline}, to control NPC behaviour and the overarching story \cite{teutenberg2013efficient} or to generate fixed, step-by-step quests such as the ones found in an MMORPG \cite{doran2010towards}.
    
 In standard systems, the story is fully scripted by a human author and presented to the player as it unfolds. This limits the capabilities for adaptation to player preferences and has very low replay value. In order to counteract this limitation, systems and frameworks have been designed to generate stories either dynamically, as it unfolds or at the start of a session \cite{brenner2010creating,teutenberg2013efficient}, by selecting story elements, ordering them and presenting them to the audience \cite{doran2011prototype, riedl2011game,yu2012sequential}. The two main aspects of narrative are the logical causal progression of a plot \cite{riedl2010narrative}, meaning that the events that occur in the narrative obey the rules of the world in which it takes place, and character believability, defined as the perception by the player that the characters act in a coherent fashion, without negatively influencing the suspension of disbelief. This means that the events in the story must appear logical and the agents must appear intentional in order for a piece of narrative to make sense \cite{dennett1989intentional}.
    
  Plans are similar to stories in that both have ordering and causality of actions in a plan sequence \cite{riedl2010narrative}.
    Prince has given a definition of narrative as “the recounting of a sequence of events that have a continuant subject and constitute a whole'' \cite{prince2003dictionary}. Therefore for a sequence of events to be considered a story, it must follow the general direction in which it started, and keep adding events that are coherent with past events.
The causality of events in a story, the relationship between the temporally ordered events that change the world state, is a property of narratives that ensures its coherence and a continuing subject \cite{chatman1993reading}. Therefore, for a sequence of events to be considered a story, it must maintain coherent causal relationships between the events. At the same time, NPC believability is dependent on the coherent causal relationships between the character's attributes known to the player, such as the character's personality or desires, and said characters' actions. This means that when deciding upon sequences of actions, characters need to make plans according to their own goals in order to appear intentional \cite{riedl2010narrative}.

As mentioned previously, plans and narratives have many things in common. Given Prince's description of a narrative \cite{prince2003dictionary}, one can see that while narratives are sequences of coherent and cohesive events that describe a series of changes in the world over the course of the story, plans consist of temporally ordered operations that transform the world state with each step, making them closely related. A planning system, given a story world and actions pertaining to said story world in the form of events, will create an artifact not unlike a story. Furthermore, partially ordered plans allow freedom in the order in which events occur if they are not causally related.
Planners require a domain theory containing all the possible events that can happen, an initial state, which, in the case of narrative planning, would be the story world and all that it contains, and a goal situation, which, in the case of narrative planning, is what the final state of the world has to be like for the goal to be considered achieved. The task of the planner is thus to find a sequence of actions that links the initial state to the goal state. With this, we are provided with an initial situation for a story, the events that unfold and the finale of the piece of narrative. 

  Although advances were made in this field, procedural generation of game narrative still has flaws. One of them is the fact that for classical planning algorithms, though they will successfully find a sequence of actions leading from the initial state to the goal state, their sequence is in no way guaranteed to make sense with the characters in the story. The planning problem itself does not concern itself with character believability \cite{riedl2010narrative}. This means that if the goal state has a princess locked up, there is nothing preventing the planning system from having the princess lock herself up and considering this a valid plan, given that it satisfies the requirements of the author. Furthermore, many systems, such as \cite{riedl2011game}, use human authored stories or human authorial intent to be able to create a coherent and cohesive story. These systems, called deliberative narrative systems \cite{riedl2010narrative}, often use centralized reasoning in order to produce a narrative that satisfies the constraints and parameters intended by the human authors, also called authorial intent. 
    
    In contrast to those systems are simulation based approaches, with a multi-agent simulation and distributed planning by and for each agent that simulate a story world. The system determines agents' actions depending on the current context and world state, solving the problem of making intentional agents \cite{teutenberg2013efficient}. By simulating a world with intentional characters, believable interactions can emerge from the simulations \cite{aylett2005fearnot}. Many systems using emergence also use director agents in order to guide the story \cite{magerko2004ai}, satisfy author goals and ensure interesting and well-structured performance of the simulation. 
    
    \subsection{Quests}
In his 2010 paper, Brenner states that plots often depend on plans failing or being thwarted and then being readjusted. Often times, in stories, multiple sub-stories or short events occur, the sum of which amounts to the overarching story. These sub-stories, in the context of games, are what we define as quests. 

\cite{aarseth2005hunt} divides quests in three basic categories. The first is \textit{place oriented}, where the player has to move their avatar through the world to reach a target location with puzzles along the way, such as in Cyan Inc.'s Myst. Slightly less common are the \textit{time-oriented} quests, where the task of the quest might simply be to survive for an interval of time. Last is the \textit{objective oriented} quest where the task is to achieve a certain objective, such as bringing an item somewhere or taking it by force from an agent in the world. These basic categories can be combined, nested and serialized to produce more or less complex quests.

The current work aims to create the proposed engine for the Creation Of Novel Adventure Narrative (CONAN) implemented in software. We will here implement and test the CONAN system on the created quests from the NPCs and the alteration of the world state as the interaction progresses. According to the theory explained above, the NPCs need to be able to provide pertinent quests to the player, the sequence of which, in a persistent world, must be coherent. Assuming that human authored quests are indeed relevant, the system must therefore be able to create similarly built quests. This experiment aims to create the quest generation engine CONAN in such a way that the quests it creates and offers to the player are similar to the ones a human author would. Furthermore, the CONAN engine is to create quests that are not only similar to the ones written by human authors but that are also relevant according to the current context of the game and the character giving out the quest in the world. We hypothesize that the CONAN system will be capable of creating quests similar to human authored quests and that these quests will be coherent, as defined by their relation to the current world state and the NPCs personalities.

\section{CONAN Design}

In this project we have created a quest-generation engine called CONAN, implemented in software. The CONAN engine's goal is to produce, given an initial state and domain files with all possible actions, novel and coherent quests for a player audience. The quests will be produced and represented as plans within the CONAN engine. Additionally, throughout quest resolution, player actions and other world-state altering events, it is able to produce more context-relevant quests as the simulation goes on, effectively producing countless different quests as the world state changes. In order for this to be possible, it requires as input a world definition composed of locations, non-player characters with pre-defined preferences, monsters and items, laws governing the world (such as "when trees are cut down, there are fewer trees"), what actions are possible, as well as the prerequisites and results of said actions. Each of these items in a specific world simulation will be objects within lists representing locations, characters, monsters or items. Each of the objects are defined within the world state by statements such as location(Castle) pertaining to the castle object, defining it as a location. These specifications will determine what is possible within the world and thus which quests can be created.

Once the input is given, the CONAN engine will accomplish its goal by having the NPCs make relevant and coherent plans to solve their goals in accordance with their preferences, which are provided in their individual domain files. These plans, which will now be referred to as quests, are to be transferred to the player character as requests. For instance, a baker NPC that ran out of bread might want to make more bread to sell, for which he needs more wheat. That NPC might then ask the player character to get him some wheat from the field in exchange for payment. 

The following section will describe in detail each element in the system.

\subsection{Initial State}

The initial state contains statements describing all that exists in the story world. We specifically use the modified Aladdin world that has seen much use in the literature \cite{teutenberg2013efficient,haslum2012narrative,riedl2010narrative} for comparison purposes. This means that the following elements are present:

\begin{itemize}
\item King Jafar who lives in a Castle
\item Aladdin, a knight who has a cooperative attitude towards King Jafar
\item Jasmine who also lives in The Castle
\item A Genie who is in the location `Magic Lamp' and unable to get out
\item A Dragon who lives in the Mountain, is hostile to agents and guards the `magic lamp'
\end{itemize}

One may notice that locations have also been mentioned in the above description. These exist in the world and are interconnected such that one may move to and from the castle and the mountain. And the Magic Lamp is only accessible from the cave. 

Each of the agents is provided preferences, which are represented by values weighting the actions, described in the domain section (see Table \ref{tab:aladdin_preferences}). 
Examples of this might be `being free' for the genie, `keep King Jafar alive' for Aladdin or `acquire wealth' for King Jafar. These statements might then all be goals for the characters for which they need to create plans to achieve. These preferences are specifically implemented in the agents as higher or lower costs to each action. Aladdin might then have a lower cost to the `Defend' action. These will be used by the planning system to find goals and sequences of actions for each agent such that they match the agent's characteristics. This is our attempt to make  their actions appear intentional.

For further testing, we also use a second initial world state. The purpose of this is to see the effect of a more complex world with more characters, locations, monsters and objects on the scalability of the system and the difference in possible quests generated by the more complex environment. This second world is set to have the following:

\begin{itemize}
\item Agents
\begin{itemize}
\item Baker
\item King
\item Lumberjack
\item Blacksmith
\item Merchant
\item Guard
\item Daughter
\end{itemize}
\item Locations:
\begin{itemize}
\item The Castle, connected to the village
\item The Village, connected to the castle, the bakery, the shop, the wheat field and the forest
\item The Wheat field, connected to the village.
\item The Cave, connected to the forest.
\item The Bakery, connected to the village
\item The Forge, connected to the cave
\item The Forest, connected to the village and the cave
\item The Shop, connected to the village
\end{itemize}
\end{itemize}

As well as several items such as a hammer, wheat, a sword and a magic spell book. The monsters (the troll, the wolves and the slimes) will also be considered as items for implementation purposes. 

\subsection{The domain file}

The domain files contain the set of possible actions that the characters in the story may use to achieve whatever goal they may have. Following the analysis presented by \cite{doran2011prototype}, the agents will have access to all the atomic actions found in his structural analysis of quests. This is so that the generated plans may include all possible actions and therefore offer the greatest variety of quests and closer resemble human created quests. The actions are as follows: DoNothing, Capture, Damage, Defend, Escort, Exchange, Experiment, Explore, Gather, Give, GoTo, Kill, Listen, Read, Repair, Report, Spy, Stealth, Take and Use.

This set of actions covers the set of possible actions that quests require players to perform in human written quests from commercial video games \cite{doran2011prototype} and will be evaluated as a set of actions for the current simulation. 

Each action will be implemented in the system the following way, using PDDL syntax, which is the standard for planning algorithms \cite{russell1995modern}:

\begin{verbatim}
(:action move
    :parameters (?p ?to ?from)
    :precondition (and (location ?to) 
        (location ?from) 
        (player ?p) (at ?p ?from))
    :effect (and (at ?p ?to) 
        (not (at ?p ?from)) 
        (increase (total-cost) 2)))
    
\end{verbatim}
Each action thus contains a name, the parameters used (player, destination and current location in the case of the above example), preconditions establishing what conditions must be true in order for the action to be available (such as the destination being a location) and what the effect of said action will have on the world once it has been performed (such as the location of the player is now the destination). Additionally, actions will be weighted in accordance with agent preferences, with actions incompatible with the agent's preference having higher cost than those in line with the preferences. This is represented with the `increase (total-cost)' part of the effect, with the number representing the agent's preference for this specific action. All actions have a base cost of 2.
This means that an action such as Kill will have a path cost higher (raise to a cost of 3) for agents such as the baker than the knight, for instance. These mappings of actions to preferences of the agents will guarantee that the agent's choice of actions will remain coherent with its personality and will protect suspension of disbelief. The action preferences of the agents are described in Table \ref{tab:aladdin_preferences}.  
\begin{table}[h!]
\centering
\caption{\label{tab:aladdin_preferences}Aladdin World character preference.}
\begin{tabular}{l|r}
Character & Preference \\\hline
Aladdin & ["+kill","-exchange","-use","+escort"] \\
Dragon & ["-damage","-take","-report","+escort","-defend"] \\
Genie & ["-kill", "-exchange", "-defend", "-read"] \\
Jasmine & ['+kill', '-spy', '-take', '-stealth'] \\
Jafar & ["-kill", "-spy", "-take", "-stealth", “move'']
\end{tabular}
\end{table}

In Table \ref{tab:aladdin_preferences}, actions preceded by a `–' have a lower cost (and are therefore preferred) and those with a + sign have a higher cost, making them less desirable. Each character has been given four or five actions with differing costs, though any number of actions can be modified in this way in order to change the agent's preference. One can see that, in doing so, it is possible to steer an agent's preference for certain actions and at the same time for certain types of quests. For instance, the reader may notice how Jafar has a preference for the actions `kill', `spy', `take', `stealth', and `move'. This will cause him to prefer plans that involve sneaky actions, murder and stealing. In the same way, one can change the preferences of an agent to make them prefer defending, trading, or any personality wanted. These preferences were arbitrarily assigned to the character, and the exact preference and value is not relevant to the actual outcome of the specific quest generation. Rather, they are used to insure different NPCs will create different quests by giving them a semblance of personality. For example, if Aladdin had +stealing, while he would not be stealing an item himself when giving the PC the quest, his preferences-defined personality make him more likely to recommend stealing an item rather than lawfully buying it or crafting it. In contrast a law-abiding citizen would recommend a significantly different quest for a similar objective.

\subsection{Goal Generation}
  The goals themselves take the form of sets of statements that must become true in the world state. These statements are a combination of predicates, such as `has' or `defended', and an object, such as an agent, an item, or a location. For example one statement in a set might be (has baker wheat), meaning that in order for the goal to be accomplished, that statement has to become true. In the current implementation of the program the possible predicates are as follows, with ?i being information, ?l a location, ?o an object, ?c a character, ?p the player, and ?cl a character at a location: 
  
\begin{verbatim}
predicates = [
        "(has ?cl ?i)",
        "(has ?p ?o)",
        "(has ?c ?o)",
        "(cooperative ?c)",
        "(at ?c ?l)",
        "(character ?c)",
        "(captive ?p ?c)",
        "(damaged ?i)",
        "(defended ?c)",
        "(defended ?i)",
        "(sneaky ?p)",
        "(dead ?m)",
        "(experimented ?i)",
        "(explored ?l)",
        "(used ?i)"]

\end{verbatim}

These define who or what can have the predicate describing a certain situation. Therefore, the player can only make herself `sneaky' as the `sneaky' predicate can only be attributed to ?p, the player.

The engine uses two algorithms to choose goals in order to compare them. The first one will randomly select goals for each NPC to use as a baseline comparison against the preference-based goals. It does so by choosing a number of random predicates from the above list equal to a user-defined number. It then cycles through the predicates and assigns a random legal item in place of the `?' object.

\begin{figure}
\centering
\includegraphics[width=0.5\textwidth]{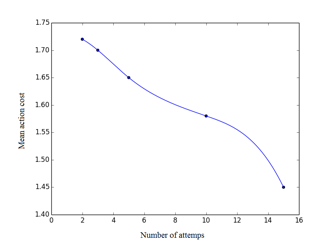}
\caption{\label{fig:meanactioncost}Mean of action cost by number of goal states attempted. It shows a downward trend of the discrepancy between the generated quest and the NPC's preferences based on the number of iterations before settling on a quest.
}
\end{figure}

The second proposed algorithm will, for each agent, select a number of random goals in the same method as was described above and rank them according to the current agent's preferences and only keep the one best fitting of the agent's preference, that is, the one with the best score. This is done by taking the list of goal states for each agent and finding a plan to reach said goal. The average cost of the actions in the plan is calculated, giving an idea of how well the plan fits with the current character's preferences. Given that the cost for actions in line with the preference of an agent is 1, the closer the average is to 1 the more the plan is in line with the character. This is repeated for each goal state and the one yielding the plan with the lowest cost is kept. The engine creates 4 random goal states to be evaluated in this fashion for each character by default. A higher number means the system is more likely to find a better goal but also means it is more taxing in terms of computation as it must create a plan for each said goal. The spline curve in Figure \ref{fig:meanactioncost} shows how the mean of action costs within a plan changes with the number of goal states attempted. This second method ensures the goals will more likely be compatible with the agents' preferences, thus preserving the illusion of intentionality. 

\subsection{Simulation}
  The simulation itself runs in a turn-based manner. There are two turns: the NPC turn and the PC turn. This means that the engine, after taking in the initial state and domain files with the actions, will solve each agent's planning problem and present the player with each plan. This is the NPC turn, where each of the agents designs their plans and gives it to the player. Before it is given to the player, a distribution of the quest motivations for the generated quests will be computed for evaluation purposes. In future versions, before selecting the quest to be given to the player, it will be compared with the distribution and will only be used if it has not been overused. This will use a player model to determine player preference and flatten the quest motivation distribution, only letting the ones the current player prefers be presented in higher frequency. This is in order to prevent the quests presented to the player from being repetitive. A translation module then transforms the plans from the series of action names to a readable form for the player to be able to understand. 
    
Next is the player turn, where the player may perform an action that impacts the game world. It is the opportunity for the player to interact with the NPC--that is, to help them with the quest they have transferred onto him. Once the action is taken, the world states are updated and the system cycles back to the NPC turn, checking for new plans in light of the changes that have occurred in the world. 

\begin{figure}
\centering
\includegraphics[width=0.35\textwidth]{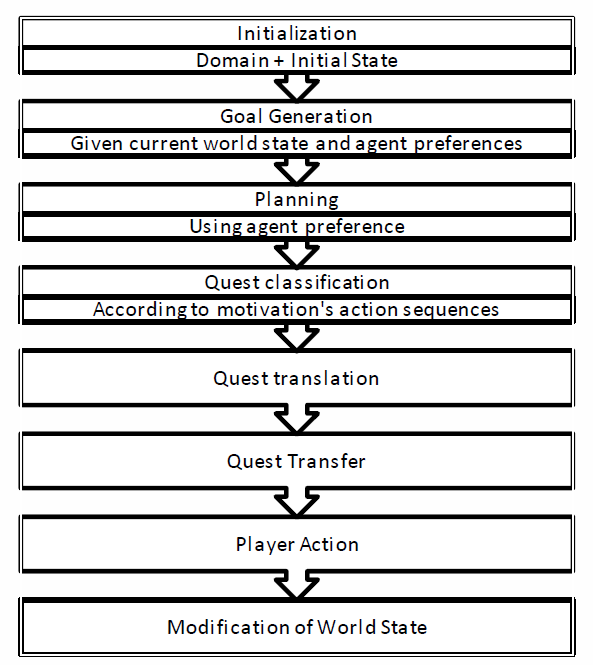}
\caption{\label{fig:flow}Flow of operation during the course of a simulation. The operation cycles from Goal Generation onwards.}
\end{figure}

\subsection{Quests}

As mentioned previously, the CONAN engine presents the player with various quests at each time step. ``Quest'' is here defined as a series of steps that must be taken in order to solve some problem, as presented by the agents to the player as a request. The quests are formally and structurally defined as follows.

In their analysis of more than 3000 human authored quests designed for commercial video games, \cite{doran2010towards} classified quests into various categories. They observed that each quest could be classified into one of 9 different categories according to its underlying motivation. Motivation is essential for ensuring that quests appear intentional and appropriate rather than randomly generated \cite{doran2011prototype}. The nine categories of motivation are as follows: knowledge, comfort, reputation, serenity, protection, conquest, wealth, ability, and equipment. The above example pertaining the baker's wheat would fall under the ``Wealth'' category. Table \ref{tab:categories} describes examples of quests that the CONAN engine could generate for each of the different categories. In order to create a system that is able to create a wide variety of realistic quests, our goal was for the engine to create quests from each of these categories. 

The example quests given below assume an initial world state as described in the second, more complex state in the Initial state section.

\begin{table}[h!]
\centering
\caption{\label{tab:categories}Categories of Quest Motivation.}
\begin{tabular}{p{15mm}|p{50mm}}
Motivation & Example of quest \\\hline
Knowledge & “Find the location of the king's stolen treasure'' \\
Comfort & “Get rid of the wolves in the forest that are preventing the lumberjack from getting wood.'' \\
Reputation & “Get granite and build a statue of me in the town square.'' \\
Serenity & “Rescue the daughter of the baker that was taken by a troll.'' \\
Protection & “Go kill the troll that has been traumatizing the village'' \\
Conquest & “Go kill my enemies.'' \\
Wealth & “Go get some wood for the lumberjack to sell.'' \\
Ability & “Find me the ancient spell book.'' \\
Equipment & “Repair the lumberjacks' axe.'' \\
\end{tabular}
\end{table}

Furthermore, given the actions that exist within the current software and its implementation, the quests will fall within the first and last categories of the tree, as described by \cite{aarseth2005hunt}, 
`place oriented' and `objective oriented.' This is because, they are to be NPC's plans to attain an objective within the game world, in different areas. The second category, time-oriented, cannot exist in the current version as there is no implemented concept of duration.

Using NPC preferences and a varied action set results in a variety of possible quests proportional to the variety of character motivation, action, and objects in the world implemented at initiation, spanning possibly the entire range of possible categories. These are the building elements of the quests and define the type of quests that will be generated by giving it its concrete goal. Smaller, simpler initial states will create fewer, less varied quests while more complex worlds will create quests spanning all quest structures.

The solution to each NPC's current problem or objective will be processed individually by a planning agent. In order to do that, each NPC is instantiated as a planning problem with its own set of constraints and goals and its own domain. As \cite{teutenberg2013efficient} have noted, although authorial intent suffers from such distributed processing, characters are more realistic, which is important for character believability, which had been noted as providing meaningful interaction \cite{aylett2005fearnot,teutenberg2013efficient}.

  The quests, as mentioned previously, will be the plans the system constructs for each agent. This means that for each agent, during the NPC phase, the system will take the current state of the world and the constraints imposed by the specific agent and find the least expensive path, in terms both of number of actions and of weight relative to its preference, to its current goal. For instance, in a world defined as follows:


The planning engine might come up with the following plan:

\begin{figure}
\centering
\includegraphics[width=0.45\textwidth]{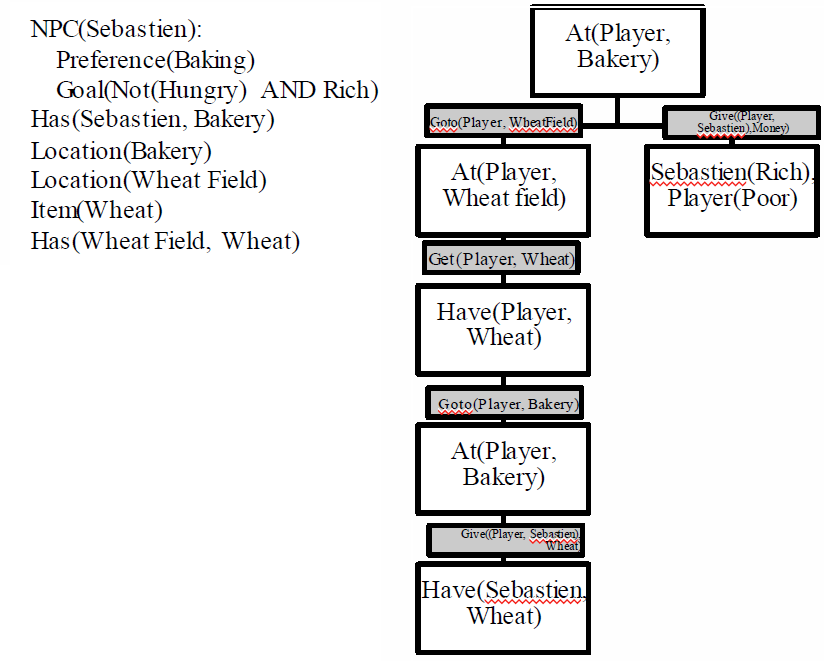}
\caption{\label{fig:sampleplan}Sample world state and plan for solving the need for wheat from the baker. Both branches show possible plans solving the goals of the NPC.}
\end{figure}

Once the agent knows the plan is possible, achieves current goal, and has the lowest cost (therefore meaning it is in line with its preferences), it is accepted as the current plan and is translated for the player to be able to read.

\subsection{Planning algorithm}
The CONAN engine need not create immensely complex plans (or quests) as the complexity of the narrative is theorized to emerge from the variety of interactions the player has with the world's agents. Furthermore, given that the planning is done by each agent for themselves, the intentionality of the plan can be dealt with directly by the planner during planning, where the preferences provide differing costs to different actions.
More complex distributed algorithms such as MAPL \cite{brenner2010creating} is unnecessarily complex for the proposed engine and the handling of intentionality by the global planner \cite{riedl2010narrative,teutenberg2013efficient} might not be relevant to the current system, as the use of weighted actions relative to character preferences might be more than enough to guarantee intentionality.
The CONAN system uses Fast Downward \cite{helmert2006fast}, as distributed under the GNU General Public License. It is an implementation of classical planning system based on heuristic search. It is encoded in PDDL 2.2, which is a standardized format for planning, and supports features such as ADL conditions and effect. This had the advantage of allowing numerical values to be used in the domain file, which we use here for weight of each action to be added in the plan weight. ADL also provides support for negative statements in preconditions. The planner supports many heuristics and we used A* in the current implementation. It takes in a standard PDDL format problem, translates it into multi-valued planning task and uses a hierarchical decomposition of the planning task to compute the heuristic function. We used this implementation of the planning engine as it supports negative clauses and numerical values. Furthermore, it has proven itself a very efficient implementation \cite{helmert2006fast}.

\section{Evaluation}
We wanted to test to see if CONAN could make the variety of quest types that have been observed in human-authored quests. To do this, we had CONAN generate a large number of quests and classified them. If all of the quest types were well-represented, then we would consider CONAN to be successful in this regard. 

The created quests are plans, which consist of sequences of actions leading the current state to the goal state. Each action sequence is compared against sequences of actions defined in \cite{doran2010towards} that belong to strategies underlying each of the motivations. For each strategy found in the plan, its associated motivation will be given a score and the motivation with the highest score will be determined as the motivation for the quest. The engine will be determined to be exhaustive in its breadth of possible quests if it is able to cover all the nine motivations that were found underlying all the human authored quests in the analysis by \cite{doran2010towards}. 

\subsection{Classifier}
  In order to determine the motivation category of each of the created quests, the engine has a built-in classification module. This module uses the quest structure from the classification of \cite{doran2010towards} to compare it against the ones it created. In their work, based on human written quests from commercial video games, they have found that all quests can be associated with one of nine broad categories, the motivations, which are shown in Table 2. This broadest classification represents different types of goals the NPC wanted achieved through the quest itself. In order to achieve the goal of the NPC, the quest employed different strategies, which are specific to said motivation, each motivation having between two and seven different strategies specific to itself, the exhaustive list of which can be found in \cite{doran2010towards}. The strategies themselves, being a way to reach the quests' goal, are high level representations of a specific sequence of action that must be performed in order to attain the goal. For example, the quest motivation “Knowledge'' has four associated strategies, which are unique to it. These are “Deliver item for study'', “Spy'', “Interview NPC'', and “Use and item in the field''. Each of these strategies is described by a sequence of action, as shown in Table \ref{tab:knowledge}.

\begin{table}[h!]
\centering
\caption{Strategies and sequences of actions for the Knowledge motivation}
\label{tab:knowledge}
\begin{tabular}{p{15mm}|p{27mm}|p{30mm}}
\textbf{Motivation} & \textbf{Strategy}        & \textbf{Sequence of action}                                                                            \\ \hline
\textbf{Knowledge}  & Deliver item for study   &    \textless get\textgreater \textless goto\textgreater give\\
                    & Spy                      &        \textless spy\textgreater  \\
                    & Interview NPC            &  \textless goto\textgreater listen\textless goto\textgreater report \\
                    & Use an item in the field & \textless get\textgreater \textless goto\textgreater use\textless goto\textgreater \\ 
                    & & \textless give\textgreater
\end{tabular}
\end{table}

The classifier, in order to determine to which broad category, or motivation, the created quest belongs to, creates a fuzzy membership based classification where each strategy belongs (0 to 1) to a quest. To do this, the quests are segmented into pairs of actions, for which the classifier searches through every strategy, for each of the motivations. If a pair is found, its score is augmented. This way, each strategy gets a score and, in turn, the sum of the scores gives the motivation its own score. The motivation with the highest score is determined to be the motivation for the quest. This can be described by the following:

$$\sum_{i}^N\frac{1}{L_{i}\times N}$$

Where $N$ is the number of strategies in that motivation, and $L_i$ is the length of a given strategy, specifically the number of pairs of actions.
This method of classifying quests has the advantage of allowing the classifier to detect strategies if they are not complete, as long as the sequence of action is closer to a given strategy than another. One inherent flaw to this process comes from the classification itself. Since some strategies are comprised of only one action, such as the `Ability' motivation with its `Use', and `Damage' strategies, the classifier will consider a sequence of actions such as (`Move' `Damage') to be both from ability because of the `Damage' and from the `Serenity' motivation, as one of its strategies is (`Move' `Damage'). This causes the motivation with single action strategies to be over represented, such as `Ability'. Furthermore, the classifier also does not account for possible sub quests or imbedded quests. For instance, if the quest was to destroy something with an axe but the character first had to get said axe, the classifier would count this as only one quest with one motivation instead of a quest with an imbedded sub quest.


In order to test the classifier itself, its output was compared to motivations assigned to quests by two of the authors. 

We took the first 50 quests that were output by the below-described large world test and classified them by hand into each of the 9 motivation categories. We then used the same 50 quests and used the classifier to divide them into the 9 motivations. Out of the 50 motivation assignments, 14 were not identified by the classifier. The inter-rater agreement between the two humans is 0.44, as measured by Krippendorff's alpha, where 1 indicates perfect reliability, 0 indicates no agreement at all, and a negative number represents systematic disagreement \cite{hayes2007answering}. The agreements for each human paired with the classifier is 0.40 and 0.38, and the agreement for the three together is 0.42. While the agreement is generally mild, the results are close to each other. This may indicate that the classification task is hard to complete, and that two humans and the classifier agreed at a similar level.

The difference in classification between the module and we can be explained in part by the inherent nature of the quest structure found in \cite{doran2010towards}. Although the motivations have specific strategies, some of them are very similar. For instance, the Knowledge motivation has the “deliver item for study'' and the Equipment motivation has the “deliver supplies'' strategy. Both these strategy have for sequence of action “\textless get\textgreater \textless goto\textgreater \textless give\textgreater ''. Such similarities lead to ambiguity in decision of which motivation should be assigned to a given quest. 

\subsection{Large World Test}


  In order to verify the breadth of possible quests, the simulation needed to create many quests. A world, simply called the Large World, was used to create about a thousand quests per non-player character. Those quests can be seen as initial quests, before any input from a player character. This allows us to investigate the diversity of quests from a single world state. Also, while a non-player character's preferences were still used to create quests (in terms of the choice of preferred steps to complete it), the choice of goals were not influenced by them. Those initial random goals allowed the engine to not be compromised by strong human-authored preferences, which could have shifted the distributions of quests.
    
\begin{figure}
\centering
\includegraphics[width=0.45\textwidth]{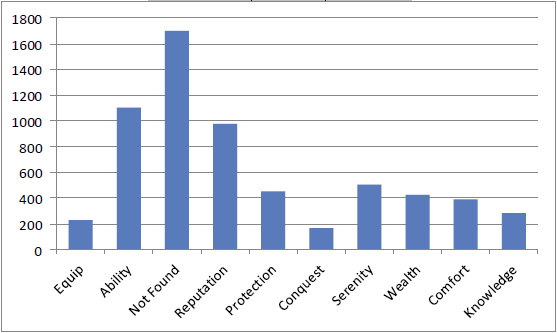}
\caption{\label{fig:motivation_large_world}Distribution of quest motivations as classified by the automatic classifier for the Large World Test. While any world with differing sets of facts will produce varying distributions, all types of quest categories are accounted for in the Large World Test.}
\end{figure}

This test can be seen as sampling the quests offered by a world. Every sampling instance, each agent tried to create a single quest. The quests were then classified using the classifier module into the 9 motivations found in \cite{doran2011prototype} depending on the strategies found in the plans. Figure \ref{fig:motivation_large_world} shows the resulting motivation distribution.

\begin{figure}[h!]
\centering
\includegraphics[width=0.45\textwidth]{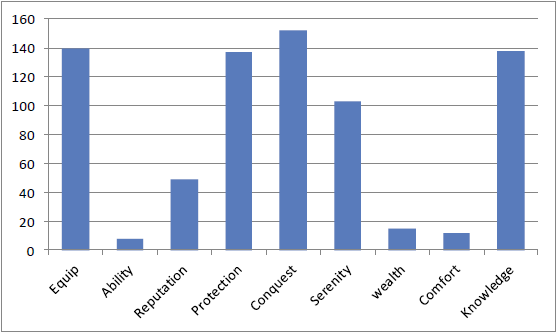}
\caption{\label{fig:human_written}Distribution of motivation in human written quests, as reported by Doran and Parberry in \cite{doran2011prototype}. While all motivations are found, the distribution is different from what we have seen in quests generated by the engine in the Large World Test}
\end{figure}

As one may see from Figure \ref{fig:motivation_large_world}, all the motivations were found in the Large World Test. Using a single initial state, the agents were able to create quests spanning all of the broad categories describing all human-writing quests. This confirms that the engine is indeed capable of creating a wide variety of quests to present to the player audience. The distribution in Figure \ref{fig:motivation_large_world} appears to be different from the distribution of human-generated quests seen in Figure \ref{fig:human_written}, as described in \cite{doran2011prototype}. One may speculate that a possible explanation for this difference can be attributed to human author preference and current popularity in the game market in the case of the human authored quests, where the engine does not have such bias towards a specific quest motivation.

Many quests fall under the `NotFound' category. These are quests that were either impossible to execute or already complete within the world. Given that for this test, the goals were chosen randomly, it was not impossible for the engine to choose goals that already exist within the world, such as (alive baker). These goals could not produce quests and were therefore placed under the `NotFound' category. Similarly, quests for which a successful plan could not be found within 5 minutes of searching were also dismissed as `NotFound'.  Any quest with the `NotFound' category would not be transferred to the player, being considered an unplayable quest. The goal generation did not automatically remove goals statements that already existed in the world state for two reasons. Firstly, although it would be beneficial for smooth gameplay, the existence of already resolved quests does not impact the capability of the engine to create other quests. Secondly, these already resolved goals could be used, in future implementation, for quests creation as well. Indeed, although the current implementation does not have this feature, goals that are already reached could be considered states that must not be changed. For instance, if the guard was to choose the goal (defended village), and such a statement already existed in the world state, one could interpret this as a quest to keep the village defended. Future implementation could use this to create even more different quests, possibly implementing the time-oriented quests from \cite{aarseth2005hunt}. As for impossible quests, these are quests where the engine tried and failed to create a plan given the goals. This could be for different reasons such as conflicting goals. 

\subsection{Aladdin World Test}

\begin{figure}
\centering
\includegraphics[width=0.45\textwidth]{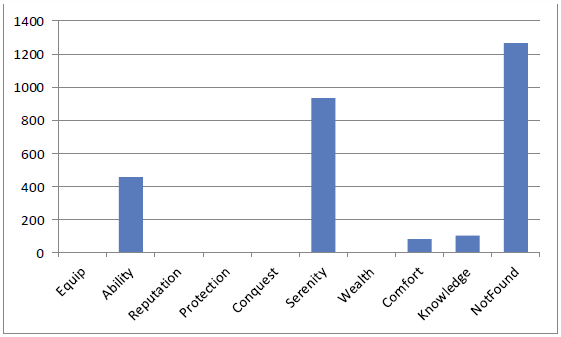}
\caption{\label{fig:aladdin_distribution}Distribution of quest motivations in Aladdin World, as classified by the automatic classifier. We observe that a more limited world, with a smaller set of facts and characters, produces a narrower range of quests in terms of motivations.}
\end{figure}

The second test used the modified Aladdin world for its simulation. This world is much simpler, having fewer facts and objects than the Large World. The same test was performed, with random goals, no player character input, and about a thousand quests for each of the 5 agents. The results are seen in Figure \ref{fig:aladdin_distribution}. Given its much simpler nature, the world did not produce the same variety of quests as the Large World.

\section{Conclusion}
  In conclusion, the CONAN engine is indeed capable of creating quests that resemble human written quests in a structural manner. Although the tests showed a different motivation distribution for the sets of quests found from the human written quests, the engine can indeed create quests that cover the entire set of categories. It has also shown that although worlds with few objects could not produce all possible quest motivation, the CONAN engine can produce them given enough information, supporting the claim that the complexity of the produced quests depends on the complexity of the world given as input. The more complex the given world, the more complex the quests presented to the player audience will be. 
  
  
  Since the generative nature depends of the world state at the time of generation, the set of possible quests at any moment is constrained only by the current world, which changes as the simulation goes on, effectively creating a constantly changing set of possible quests.
    
  Such an engine, effectively capable of autonomously creating infinity of quests would be valuable to an industry that is growing in demand for complexity and which uses human artists and authors to create such content. The possibility to offload this task, partially or totally onto an autonomous system would make the creation of such games much cheaper and create a very high replay value as the gameplay would be different every time the game is played by its intended audience.
    
  The current work differs significantly from other similar systems by its reliance on emergence and player interactions. State of the art systems such as IMPRACTical \cite{teutenberg2013efficient}, seek to create stories through intentional planning by multiple agents and a single narrative planner that generates the narrative story.  \cite{brenner2010creating} uses continuous multi-agent planning in order to write a story with its different agent's goals and intentions, and \cite{riedl2011game} uses a centralized planner to adapt plotlines in order to create new stories. Lastly, \cite{doran2010towards} use structural rules to create quests from their analysis. The CONAN system's main difference comes from its simulation based, emergent approach and its iterative process where story is generated through each iteration of the player's actions. 
  
  We forgo the use of story planners or director agents in favor of an interactive approach where the player dictates, through their decision during the simulation, how the story ought to evolve. This approach has yet to be investigated in depth and can provide a different approach to generation of stories in games, which would be less involved in terms of human authoring and thus more efficient than the traditional human written stories. 
  
It could be that a coherent story will emerge from such an engine given a simulation-based system with believable characters that provide the human player with believable interactions in the form of quests. The human player, through their interactions in the game, might act as their own director agent, forming a coherent narrative through their own choice of quests. 

In director agent systems \cite{magerko2004ai,laird2001human}, the AI entity chooses interesting plot points and adjusts the story to maximize its entertainment value. We suggest here that the player's choice of which quest to undertake within a large set of available quests, which sub-story to pursue, and which actions to perform in the world will ensure that the story elements, the quests as they unfold, will be interesting to the audience, the human player.

Future work should investigate the interaction aspect of this idea. Specifically, one might look at subjective experience of player reporting on both their interactions with the characters in the world but also of their experience of any single simulation's story as they progress through the quest and interact with the agents in the world, giving information on the believability of the characters. 

Furthermore, studies should investigate the current engine capabilities for emergence of story. The current study does not assess believability of the agents nor does it investigate the effect of the simulation when presented to, and interacting with a human player. This interaction is theorized to create, in the mind of the audience, a story that is personal and different for every player, every time. Furthermore, this study does not investigate qualitative and subjective assessment of interestingness or suspension of disbelief. Studies looking to further this engine should implement it into a proper game and test it using participants.


%




\ifCLASSOPTIONcaptionsoff
  \newpage
\fi



\bibliographystyle{IEEEtran}
\bibliography{no.bib}

\end{document}